\documentclass[a4paper]{article}
\pdfoutput=1
\usepackage{graphicx}
\usepackage{amsmath,amssymb}
\usepackage{color}
\usepackage{epsfig}
\usepackage{graphicx}
\usepackage{amsmath}
\usepackage{amssymb}
\usepackage{color}
\usepackage{mdwlist}
\usepackage[all]{xy}
\usepackage{algorithm}
\usepackage{algorithmic}
\usepackage{eucal}

\newcommand{\inner}[2]{\left\langle #1,#2 \right\rangle}
\newcommand{\nbr}[1]{\left\|#1\right\|}

\DeclareMathOperator*{\argmax}{\mathrm{argmax}}
\DeclareMathOperator*{\argmin}{\mathrm{argmin}}

\usepackage{theorem}
\usepackage{array}
\usepackage[pagebackref=true,breaklinks=true,letterpaper=true,colorlinks,bookmarks=false]{hyperref}

\newcommand{\eq}[1]{(eq.~\ref{#1})}
\newcommand{\xhdr}[1]{\vspace{1.7mm}\noindent{{\bf #1.}}}

\begin{document}


\title{Image Labeling on a Network:\\Using Social-Network Metadata for Image Classification}
\author{Julian McAuley and Jure Leskovec\\Stanford University\\{\tt\small jmcauley@cs.stanford.edu, jure@cs.stanford.edu}}

\maketitle

\begin{abstract}
Large-scale image retrieval benchmarks invariably consist of images from the Web. Many of these benchmarks are derived from online photo sharing networks, like Flickr, which in addition to hosting images also provide a highly interactive social community. Such communities generate rich metadata that can naturally be harnessed for image classification and retrieval. Here we study four popular benchmark datasets, extending them with social-network metadata, such as the groups to which each image belongs, the comment thread associated with the image, who uploaded it, their location, and their network of friends. Since these types of data are inherently relational, we propose a model that explicitly accounts for the interdependencies between images sharing common properties. We model the task as a binary labeling problem on a network, and use structured learning techniques to learn model parameters. We find that social-network metadata are useful in a variety of classification tasks, in many cases outperforming methods based on image content.
\end{abstract}

\section{Introduction}

\begin{figure}
 \begin{center}
  \includegraphics[scale=0.5]{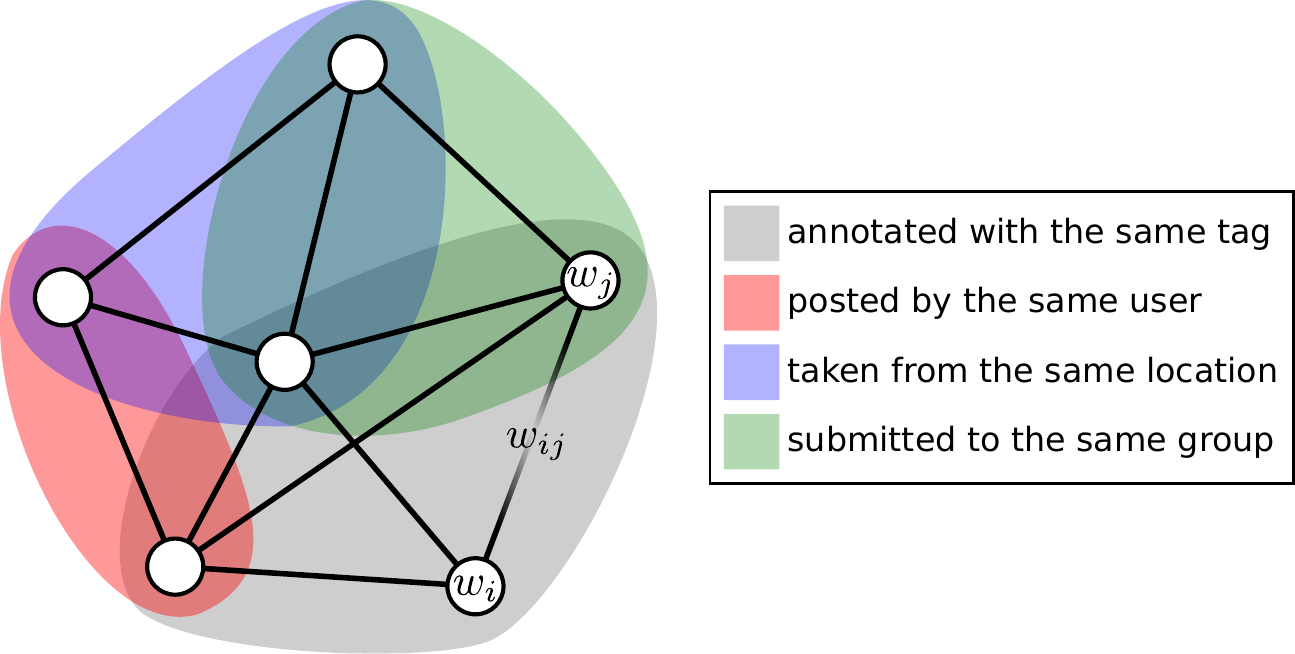}
 \end{center}

\vspace{-3mm}

\caption{The proposed relational model for image classification. Each node represents an image, with cliques formed from images sharing common properties. `Common properties' can include (for example) \emph{communities}, e.g.~images submitted to a group; \emph{collections}, e.g.~sets created by a user; \emph{annotations}, e.g.~tag data; and \emph{user data}, e.g.~the photo's uploader and their network of friends.}
\label{fig:model}
\end{figure}

Recently, research on image retrieval and classification has focused on large image databases collected from the Web. Many of these datasets are built from online photo sharing communities such as Flickr \cite{Everingham10,huiskes08,nowak10,nus-wide} and even collections built from image search engines \cite{Deng2009} consist largely of Flickr images. 

Such communities generate vast amounts of metadata as users interact with their images, and with each other, though only a fraction of such data are used by the research community. The most commonly used form of metadata considered in multimodal classification settings is the set of \emph{tags} associated with each image. In \cite{guillaumin2010multimodal} the authors study the relationship between tags and manual annotations, with the goal of recovering annotations using a combination of tags and image content. The problem of recommending tags was studied in \cite{lind}, where possible tags were obtained from similar images and similar users. The same problem was studied in \cite{sigwww}, who exploit the relationships between tags to suggest future tags based on existing ones. Friendship information between users was studied for tag recommendation in \cite{sawant}, and in \cite{facebookTagging} for the case of Facebook.

Another commonly used source of metadata comes directly from the camera, in the form of \emph{exif} and \emph{GPS} data \cite{luo06,li09,kalo09,joshi11}. Such metadata can be used to determine whether two photos were taken by the same person, or from the same location, which provides an informative signal for certain image categories.

Our goal in this paper is to assess what \emph{other} types of metadata may be beneficial, including the groups, galleries, and collections in which each image was stored, the text descriptions and comment threads associated with each image, and user profile information including their location and their network of friends. In particular, we focus on the following three questions:  (1) How can we effectively model relational data generated by the social-network? (2) How can such metadata be harnessed for image classification and labeling? (3) What types of metadata are useful for different image labeling tasks?

Focusing on the first question we build on the intuition that images sharing similar tags and appearance are likely to have similar labels \cite{huiskes08}. In the case of image tags, simple nearest-neighbor type methods have been proposed to `propagate' annotations between similar images \cite{mensink}. However, unlike image labels and tags -- which are categorical -- much of the metadata derived from social networks is inherently \emph{relational}, such as collections of images posted by a user or submitted to a certain group, or the networks of contacts among users. We argue that to appropriately leverage these types of data requires us to \emph{explicitly} model the relationships between images, an argument also made in \cite{ranking}.

To address the relational nature of social-network data, we propose a graphical model that treats image classification as a problem of simultaneously predicting binary labels for a network of photos. Figure \ref{fig:model} illustrates our model: nodes represent images, and edges represent \emph{relationships} between images. Our intuition that images sharing common properties are likely to share labels allows us to exploit techniques from supermodular optimization, allowing us to efficiently make binary predictions on all images simultaneously \cite{graphcuts}.

In the following sections, we study the extent to which categorical predictions about images can be made using social-network metadata. We first describe how we augment four popular datasets with a variety of metadata from Flickr. We then consider three image labeling tasks. The creators of these datasets obtained labels through crowdsourcing and from the Flickr user community. Labels range from objective, everyday categories such as `person' or `bicycle', to subjective concepts such as `happy' and `boring'.

We show that social-network metadata reliably provide \emph{context} not contained in the image itself. Metadata based on common galleries, image locations, and the author of the image tend to be the most informative in a range classification scenarios. Moreover, we show that the proposed relational model outperforms a `flat' SVM-like model, which means that it is essential to model the relationships between images in order to exploit these social-network features.

\section{Dataset Construction and Description}
\label{sec:datasets}

We study four popular datasets that have groundtruth provided by human annotators. Because each of these datasets consists entirely of images from Flickr, we can enrich them with social network metadata, using Flickr's publicly available API. The four image collections we consider are described below:
\begin{itemize*}
 \item The \emph{PASCAL Visual Object Challenge} (`PASCAL') consists of over 12,000 images collected since 2007, with additional images added each year \cite{Everingham10}. Flickr sources are available only for training images, and for the test images from 2007. Flickr sources were available for 11,197 images in total.
 \item The \emph{MIR Flickr Retrieval Evaluation} (`MIR') consists of one million images, 25,000 of which have been annotated \cite{huiskes08}. Flickr sources were available for 15,203 of the annotated images.
 \item The \emph{ImageCLEF Annotation Task} (`CLEF') uses a subset of 18,000 images from the MIR dataset, though the correspondence is provided only for 8,000 training images \cite{nowak10}. Flickr sources were available for 4,807 images.
 \item The \emph{NUS Web Image Database} (`NUS') consists of approximately 270,000 images \cite{nus-wide}. Flickr sources are available for all images.
\end{itemize*}
Flickr sources for the above photos were provided by the dataset creators. Using Flickr's API we obtained the following metadata for each photo in the above datasets:
\begin{itemize*}
 \item The photo itself
 \item Photo data, including the photo's title, description, location, timestamp, viewcount, upload date, etc.
 \item User information, including the uploader's name, username, location, their network of contacts, etc.
 \item Photo tags, and the user who provided each tag
 \item Groups to which the image was submitted (only the uploader can submit a photo to a group)
 \item Collections (or sets) in which the photo was included (users create collections from their own photos)
 \item Galleries in which the photo was included (a single user creates a gallery only from \emph{other} users' photos)
 \item Comment threads for each photo
\end{itemize*}
We only consider images from the above datasets where \emph{all} of the above data was available, which represents about 90\% of the images for which the original Flickr source was available (to be clear, we include images where this data is \emph{absent}, such as images with no tags, but not where it is \emph{missing}, i.e., where an API call fails, presumably due to the photo having been deleted from Flickr). Properties of the data we obtained are shown in Table \ref{tab:stats}. Note in particular that the \emph{ratios} in Table \ref{tab:stats} are not uniform across datasets, for example the NUS dataset favors `popular' photos that are highly tagged, submitted to many groups, and highly commented on; in fact all types of metadata are more common in images from NUS than for other datasets. The opposite is true for PASCAL, which has the least metadata per photo, which could be explained by the fact that certain features (such as galleries) did not exist on Flickr when most of the dataset was created. Details about these datasets can be found in \cite{Everingham10,huiskes08,nowak10,nus-wide}.

\begin{table}[t]
\caption{Dataset statistics. The statistics reveal large differences between the datasets, for instance images in MIR have more tags and comments than images in PASCAL, presumably due to MIR's bias towards `interesting' images \cite{huiskes08}; few images in PASCAL belong to galleries, owing to the fact that most of the dataset was collected before this feature was introduced in 2009.
Note that the number of tags per image is typically slightly higher than what is reported in \cite{huiskes08,nowak10,nus-wide}, as there may be additional tags that appeared in Flickr since the datasets were originally created.
}
\begin{center}
\footnotesize{
\setlength{\tabcolsep}{3.5pt}
  \begin{tabular}{|lrrrrr|}
\hline
                    & CLEF          & PASCAL         & MIR            & NUS             & ALL\\
Number of photos    & \textbf{4546} & \textbf{10189} & \textbf{14460} & \textbf{244762} & \textbf{268587}\\
Number of users     & 2663          & 8698           & 5661           & 48870           & 58522\\
Photos per user     & 1.71          & 1.17           & 2.55           & 5.01            & 4.59\\
Number of tags      & 21192         & 27250          & 51040          & 422364          & 450003\\
Tags per photo      & 10.07         & 7.17           & 10.24          & 19.31           & 18.36\\
Number of groups    & 10575         & 6951           & 21894          & 95358           & 98659\\
Groups per photo    & 5.09          & 1.80           & 5.28           & 12.56           & 11.77\\
Number of comments  & 77837         & 16669          & 248803         & 9837732         & 10071439\\
Comments per photo  & 17.12         & 1.64           & 17.21          & 40.19           & 37.50\\
Number of sets      & 6066          & 8070           & 15854          & 165039          & 182734\\
Sets per photo      & 1.71          & 0.87           & 1.72           & 1.95            & 1.90\\
Number of galleries & 1026          & 155            & 3728           & 100189          & 102116\\
Galleries per photo & 0.23          & 0.02           & 0.27           & 0.67            & 0.62\\
Number of locations & 1007          & 1222           & 2755           & 22106           & 23745\\
Number of labels    & \textbf{99}   & \textbf{20}    & \textbf{14}    & \textbf{81}     & \textbf{214}\\
Labels per photo    & 11.81         & 1.95           & 0.93           & 1.89            & 2.04\\
\hline
  \end{tabular}
}
 \end{center}
\label{tab:stats}
\end{table}

\begin{figure*}[t]
 \begin{center}
  \includegraphics[scale=0.4]{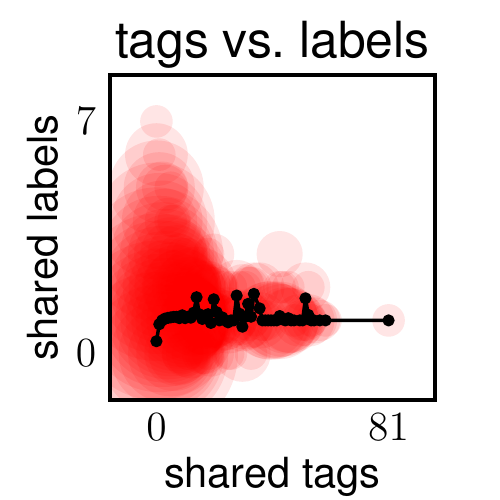}%
  \includegraphics[scale=0.4]{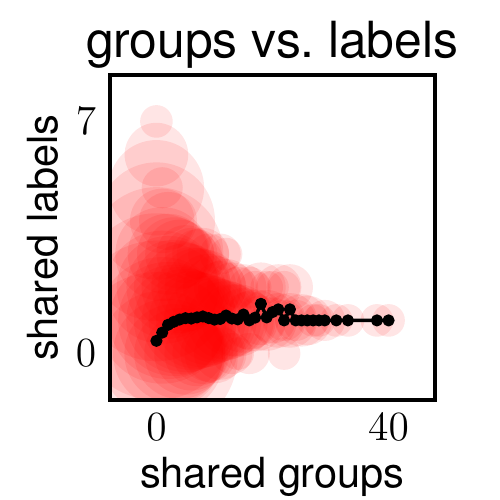}%
  \includegraphics[scale=0.4]{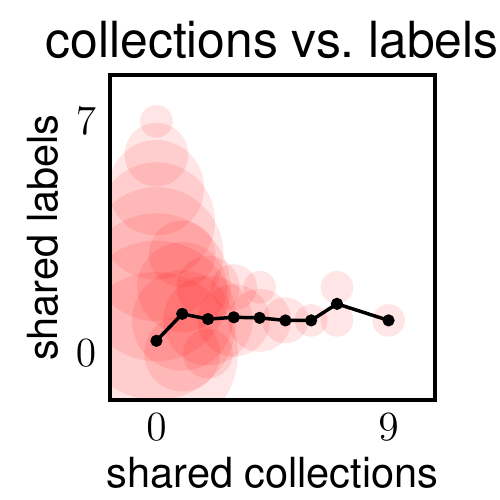}%
  \includegraphics[scale=0.4]{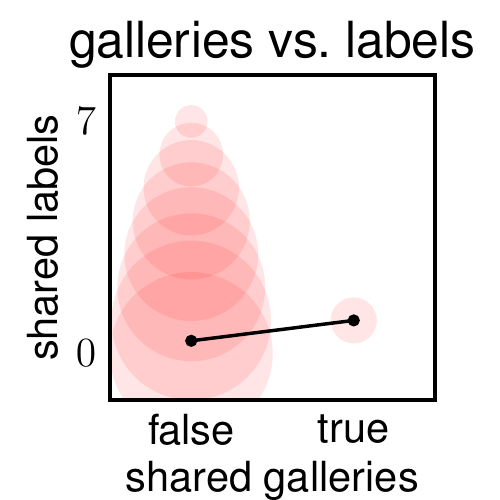}%
  \includegraphics[scale=0.4]{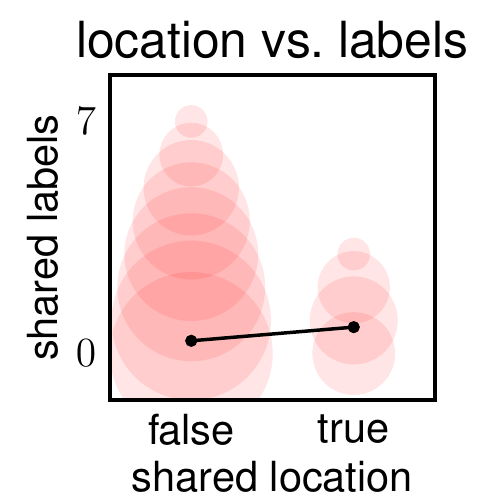}%
  \includegraphics[scale=0.4]{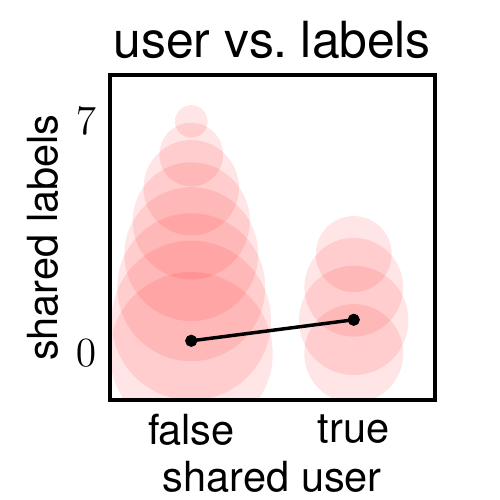}

  \includegraphics[scale=0.4]{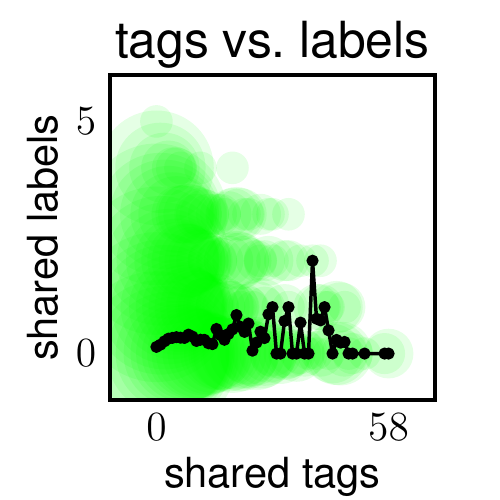}%
  \includegraphics[scale=0.4]{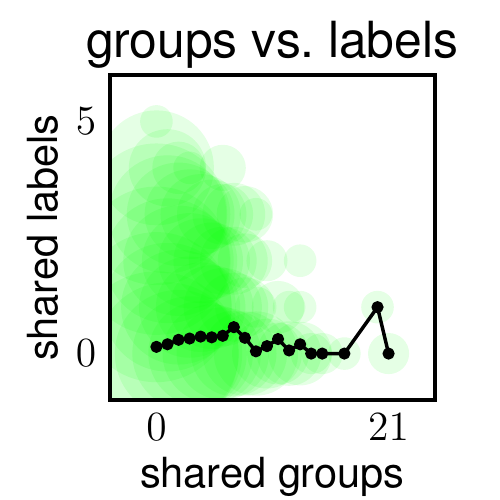}%
  \includegraphics[scale=0.4]{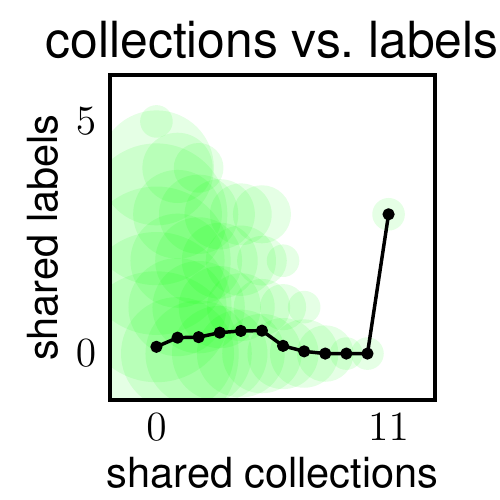}%
  \includegraphics[scale=0.4]{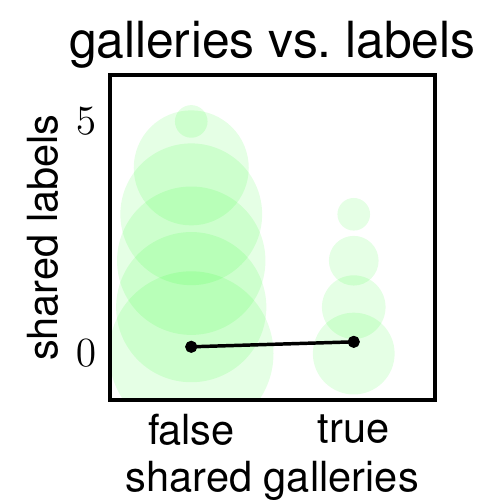}%
  \includegraphics[scale=0.4]{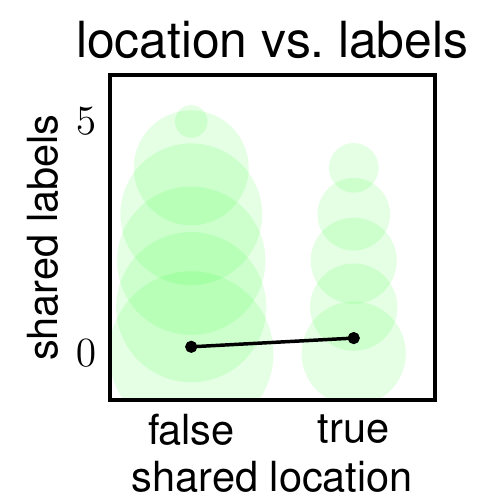}%
  \includegraphics[scale=0.4]{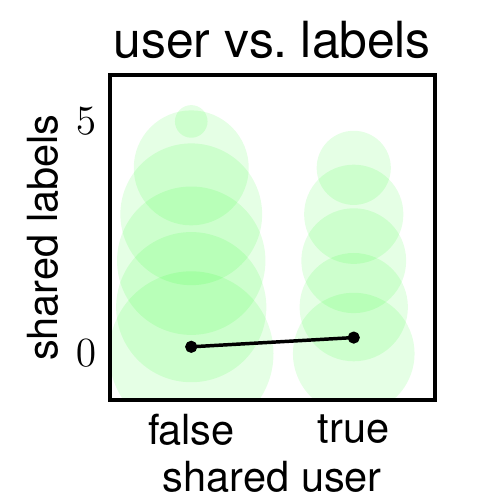}

  \includegraphics[scale=0.4]{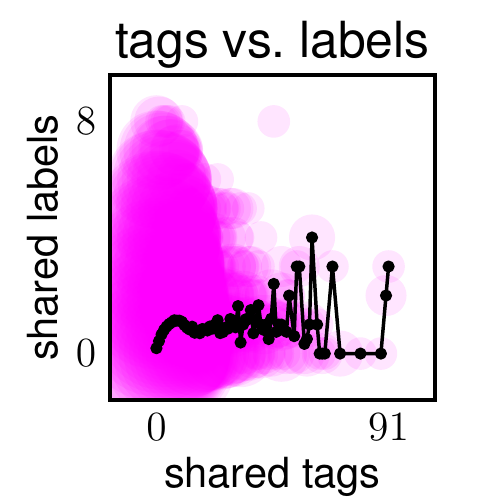}%
  \includegraphics[scale=0.4]{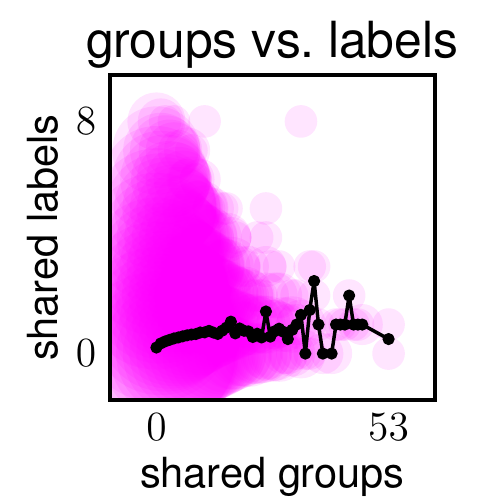}%
  \includegraphics[scale=0.4]{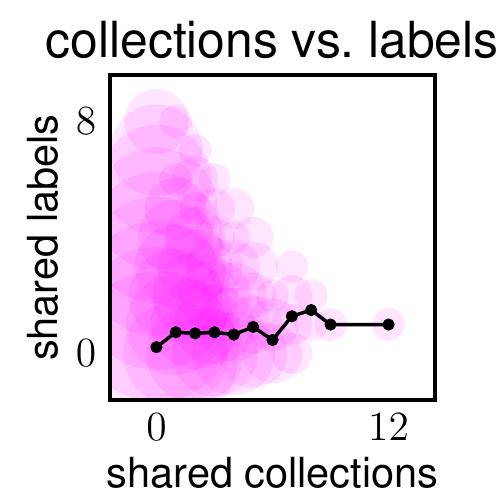}%
  \includegraphics[scale=0.4]{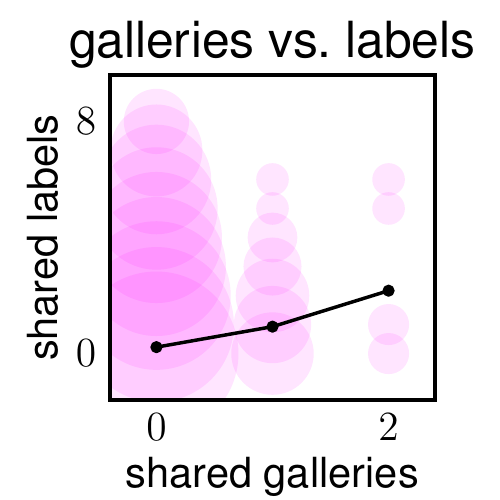}%
  \includegraphics[scale=0.4]{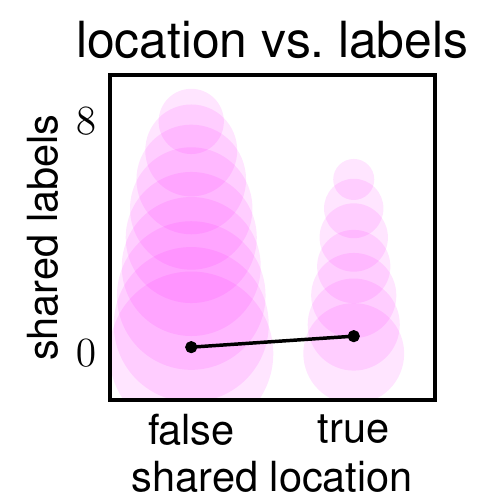}%
  \includegraphics[scale=0.4]{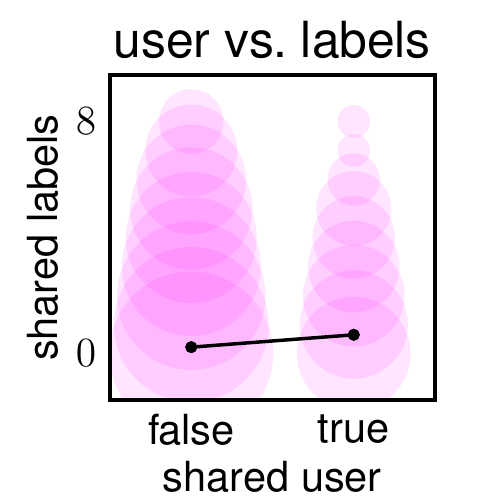}

\includegraphics[scale=0.4]{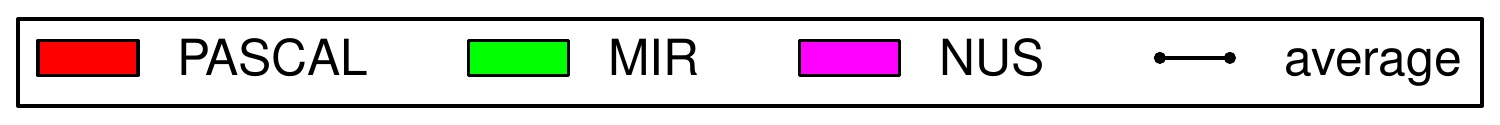}

\vspace{-5mm}
 \end{center}
\caption{Relationships between Flickr metadata and image labels provided by external evaluators. All figures are best viewed in color. Scatterplots show the number of images that share a pair of properties in common, with radii scaled according to the logarithm of the number of images at each coordinate. All pairs of properties have positive correlation coefficients. ImageCLEF data is suppressed, as it is a subset of MIR and has similar behavior.}
\label{fig:statplots}
\end{figure*}

In Figure \ref{fig:statplots} we study the relationship between various types of Flickr metadata and image labels. Images sharing common tags are likely to share common labels \cite{mensink}, though Figure \ref{fig:statplots} reveals similar behavior for nearly all types of metadata. Groups are similar to tags in quantity and behavior: images that share even a single group or tag are much more likely to have common labels, and for images sharing \emph{many} groups or tags, it is very unlikely that they will not share at least one label. The same observation holds for collections and galleries, though it is rarer that photos have these properties in common. Photos taken at the same location, or by the same user also have a significantly increased likelihood of sharing labels \cite{luo06}. Overall, this indicates that the image metadata provided by the interactions of the Flickr photo-sharing community correlates with image labels that are provided by the external human evaluators.

All code and data is available from the authors' webpages.\footnote{\url{http://snap.stanford.edu/}, \url{http://i.stanford.edu/~julian/}}

\section{Model}
\label{sec:model}

\setlength{\tabcolsep}{2.5pt}
\begin{table}[t]
  \caption{Notation}
\footnotesize{
\begin{center}
\begin{tabular}{l | p{0.67\columnwidth} }
 \hline
\textbf{Notation} & \textbf{Description}\\
\hline
\hline
$\mathcal X = \lbrace x_n \ldots x_N \rbrace$ & An image dataset consisting of $N$ images\\
$\mathcal L = \lbrace -1, 1 \rbrace^L$ & A label space consisting of $L$ categories.\\
$y^n \in \mathcal L$ & The groundtruth labeling for the image $x_n$.\\
$y^n_c \in \lbrace -1, 1 \rbrace$ & The groundtruth for a particular category $c$.\\
$Y_c \in \lbrace -1, 1 \rbrace^N$ & The groundtruth for the entire dataset for category $c$.\\
$\bar{y}_c(x_n; \Theta_c) \in \lbrace -1, 1 \rbrace$ & The prediction made image $x_n$ and category $c$.\\
$\bar{Y}_c(\mathcal X; \Theta_c) \in \lbrace -1, 1 \rbrace^N$ & Predictions across the entire dataset for category $c$.\\
$\theta_c^{\text{node}} \in \mathbb{R}^{F_1}$ & Parameters of first-order features for category $c$.\\
$\theta_c^{\text{edge}} \in \mathbb{R}^{F_2}$ & Parameters of second-order features for category $c$.\\
$\Theta_c = (\theta_c^\text{node}; \theta_c^\text{edge})$ & Full parameter vector for category $c$.\\
$\phi_c(x_i) \in \mathbb{R}^{F_1}$ & Features of the image $x_i$ for category $c$.\\
$\phi_c(x_i, x_j) \in \mathbb{R}^{F_2}$ & Features of the pair of images $(x_i, x_j)$ for category $c$.\\
$\Phi_c(\mathcal X, Y) \in \mathbb{R}^{F_1 + F_2}$ & Aggregate features for labeling the entire dataset $\mathcal X$ as $Y \in \lbrace -1, 1 \rbrace^N$ for category $c$.\\
$\Delta(Y, Y_c) \in \mathbb{R}_+$ & The error induced by making the prediction $Y$ when the correct labeling is $Y_c$.\\
\hline
\end{tabular}
\end{center}
}
  \label{tab:notation}
\end{table}

The three tasks we shall study are label prediction (i.e., predicting groundtruth labels using image metadata), tag prediction, and group recommendation. As we shall see, each of these tasks can be thought of as a problem of predicting \emph{binary} labels for each of the images in our datasets. 

Briefly, our goal in this section is to describe a binary graphical model for each image category (which might be a label, tag, or group), as depicted in Figure \ref{fig:model}. Each node represents an image; the weight $w_i$ encodes the potential for a node to belong to the category in question, given its features; the weights $w_{ij}$ encode the potential for two images to have the \emph{same} prediction for that category. We first describe the `standard' SVM model, and then describe how we extend it to include relational features.

The notation we use throughout the paper is summarized in Table \ref{tab:notation}. Suppose we have a set of images $\mathcal X = \lbrace x_n \ldots x_N \rbrace$, each of which has an associated groundtruth labeling $y^n \in \lbrace -1, 1 \rbrace^L$, where each $y^n_c$ indicates positive or negative membership to a particular category $c \in \lbrace 1 \ldots L \rbrace$. Our goal is to learn a classifier that predicts $y^n_c$ from (some features of) the image $x_n$.

\xhdr{The `Standard' Setting} Max-margin SVM training assumes a classifier of the form
\begin{equation}
 \bar{y}_c(x_n, \Theta_c) = \argmax_{y \in \lbrace -1, 1 \rbrace} y \cdot \langle \phi_c(x_n), \Theta_c \rangle,
\label{eq:svm}
\end{equation}
so that $x_n$ has a positive label whenever $\langle \phi_c(x_n), \Theta_c \rangle$ is positive. $\phi_c(x_n)$ is a \emph{feature vector} associated with the image $x_n$ for category $c$, and $\Theta_c$ is a \emph{parameter vector}, which is selected so that the predictions made by the classifier of \eq{eq:svm} match the groundtruth labeling. Note that a \emph{different} parameter vector $\Theta_c$ is learned for each category $c$, i.e., the model makes the assumption that the labels for each category are independent.

Models similar to that of \eq{eq:svm} (which we refer to as `flat' models since they consider each image independently and thus ignore relationships between images) are routinely applied to classification based on image features \cite{chapelle}, and have also been used for classification based on image tags, where as features one can simply create indicator vectors encoding the presence or absence of each tag \cite{huiskes08}. In practice this means that for each tag one learns its influence on the presence of each label. For image tags, this approach seems well motivated, since tags are \emph{categorical} attributes. What this also means is that the tag vocabulary -- though large -- ought to grow sublinearly with the number of photos (see Table \ref{tab:stats}), meaning that a more accurate model of each tag can be learned as the dataset grows. Based on the same reasoning, we encode group and text information (from image titles, descriptions, and comments) in a similar way.

\xhdr{Modeling Relational Metadata} Other types of metadata are more naturally treated as \emph{relational}, such as the network of contacts between Flickr users. Moreover, as we observed in Table \ref{tab:stats}, even for the largest datasets we only observe a very small number of photos per user, gallery, or collection. This means it would not be practical to learn a separate `flat' model  for each category. However, as we saw in Figure \ref{fig:statplots}, it may still be worthwhile to model the fact that photos from the same gallery are likely to have similar labels (similarly for users, locations, collections, and contacts between users). 

We aim to learn \emph{shared} parameters for these features. Rather than learning the extent to which membership to a particular collection (resp.~gallery, user) influences the presence of a particular label, we learn the extent to which a pair of images that belong to the \emph{same} gallery are likely to have \emph{the same} label. In terms of graphical models, this means that we form a \emph{clique} from photos sharing common metadata (as depicted in Figure \ref{fig:model}).

These relationships between images mean that classification can no longer be performed independently for each image as in \eq{eq:svm}. Instead, our predictor $\bar{Y}_c(\mathcal X, \Theta_c)$ labels the entire dataset at once, and takes the form
\small
\begin{equation}
  \bar{Y}_c(\mathcal X, \Theta_c) = \argmax_{Y \in \lbrace -1, 1 \rbrace^N} \sum_{i=1}^N y_i \cdot \underbrace{\langle \phi_c(x_i), \theta^\text{node}_c \rangle}_{w_i} +
\sum_{i,j = 1}^N\sum_{j = 1}^N \delta(y_i = y_j) \underbrace{\langle \phi_c(x_i, x_j), \theta^\text{edge}_c \rangle}_{w_{ij}},
\label{eq:gmodel}
\end{equation}
\normalsize
where $\phi_c(x_i,x_j)$ is a feature vector encoding the relationship between images $x_i$ and $x_j$, and $\delta(y_i = y_j)$ is an indicator that takes the value 1 when we make the same binary prediction for both images $x_i$ and $x_j$. The first term of \eq{eq:gmodel} is essentially the same as \eq{eq:svm}, while the second term encodes relationships between images. Note that \eq{eq:gmodel} is linear in $\Theta_c = (\theta^\text{node}_c; \theta^\text{edge}_c)$, i.e., it can be rewritten as
\begin{equation}
 \bar{Y}_c(\mathcal X, \Theta_c) = \argmax_{Y \in \lbrace -1, 1 \rbrace^N} \langle \Phi_c(\mathcal X, Y), \Theta_c \rangle.
\end{equation}

Since \eq{eq:gmodel} is a binary optimization problem consisting of pairwise terms, we can cast it as \emph{maximum a posteriori} (MAP) inference in a graphical model, where each node corresponds to an image, and edges are formed between images that have some property in common.

Despite the large maximal clique size of the graph in question, we note that MAP inference in a pairwise, binary graphical model is tractable so long as the pairwise term is \emph{supermodular}, in which case the problem can be solved using graph-cuts \cite{graphcuts,Boros}. A pairwise potential $f(y_i,y_j)$ is said to be supermodular if
\begin{equation}
 f(-1,-1) + f(1,1) \geq f(-1,1) + f(1,-1),
\end{equation}
which in terms of \eq{eq:gmodel} is satisfied so long as
\begin{equation}
 \langle \phi_c(x_i, x_j), \theta^\text{edge}_c \rangle \geq 0.
\label{eq:uality}
\end{equation}
Assuming positive features $\phi_c(x_i, x_j)$, a sufficient (but not necessary) condition to satisfy \eq{eq:uality} is $\theta^\text{edge}_c \geq \mathbf{0}$, which in practice is what we shall enforce when we learn the optimal parameters $\Theta_c = (\theta^\text{node}_c; \theta^\text{edge}_c)$. Note that this is a particularly weak assumption: all we are saying is that photos sharing common properties are \emph{more likely} to have similar labels than different ones. The plots in Figure \ref{fig:statplots} appear to support this assumption.

We solve \eq{eq:gmodel} using the graph-cuts software of \cite{Boykov01anexperimental}. For the largest dataset we consider (NUS), inference using the proposed model takes around $10$ seconds on a standard desktop machine, i.e., less than $10^{-4}$ seconds per image. During the parameter learning phase, which we discuss next, memory is a more significant concern, since for practical purposes we store all feature vectors in memory simultaneously. Where this presented an issue, we retained only those edge features with the most non-zero entries up to the memory limit of our machine. Addressing this shortcoming using recent work on distributed graph-cuts remains an avenue for future study \cite{parallelcuts}.

\section{Parameter Learning}
\label{sec:learning}

In this section we describe how popular \emph{structured learning} techniques can be used to find model parameter values $\Theta_c$ so that the predictions made by \eq{eq:gmodel} are consistent with those of the groundtruth $Y_c$. We assume an estimator based on the principle of regularized risk minimization \cite{tsoch05}, i.e., the optimal parameter vector $\Theta_c^*$ satisfies
\begin{align}
\Theta_c^*=\argmin_{\Theta} \Biggl[ \underbrace{\vphantom{\frac{\lambda}{2}}\Delta(\bar{Y}(\mathcal X;\Theta),Y_c)}_{\text{empirical risk}} + \underbrace{\frac{\lambda}{2} \nbr{\Theta}^2}_{\text{regularizer}}\Biggr],
\label{eq:opt_problem}
\end{align}
where $\Delta(\bar{Y}(\mathcal X;\Theta),Y_c)$ is some \emph{loss function} encoding the error induced by predicting the labels $\bar{Y}(\mathcal X;\Theta)$ when the correct labels are $Y_c$, and $\lambda$ is a hyperparameter controlling the importance of the regularizer.

We use an analogous approach to that of SVMs \cite{tsoch05}, by optimizing a convex upper bound on the structured loss of \eq{eq:opt_problem}. The resulting optimization problem is
\begin{subequations}
\label{eq:relaxation}
\begin{align}
\label{eq:objective}
& [\Theta^*,\xi^*]=\argmin_{\Theta,\xi} \left[\xi + \lambda \nbr{\Theta}^2\right]\\
\label{eq:constraints}
\text{s.t.~} & \inner{\Phi(\mathcal X,Y_c)}{\Theta}-\inner{\Phi(\mathcal X,Y)}{\Theta}\ge \Delta(Y,Y_c)-\xi,\\
\nonumber
& \theta_c^\text{edge} \geq \mathbf{0} \quad \forall Y\in \lbrace -1, 1 \rbrace^N.
\end{align}
\end{subequations}
Note the presence of the additional constraint $\theta_c^\text{edge} \geq \mathbf{0}$, which enforces that \eq{eq:gmodel} is supermodular (which is required for efficient inference).

The principal difficulty in optimizing \eq{eq:objective} lies in the fact that \eq{eq:constraints} includes exponentially many constraints -- one for every \emph{possible} output $Y \in \lbrace -1, 1 \rbrace^N$ (i.e., two possibilities for every image in the dataset). To circumvent this, \cite{tsoch05} proposes a constraint generation strategy, including at each iteration the constraint that induces the largest value of the slack $\xi$. Finding this constraint requires us to solve
\begin{equation}
 \hat{Y}_c(\mathcal X; \Theta_c) = \argmax_{Y \in \lbrace -1, 1 \rbrace^N} \langle \Phi_c(\mathcal X, Y), \Theta_c \rangle + \Delta(Y, Y_c),
\label{eq:colgen}
\end{equation}
which we note is tractable so long as $\Delta(Y, Y_c)$ is also a supermodular function of $Y$, in which case we can solve \eq{eq:colgen} using the same approach we used to solve \eq{eq:gmodel}. Note that since we are interested in making simultaneous binary predictions for the entire dataset (rather than \emph{ranking}), a loss such as the average precision is not appropriate for this task. Instead we optimize the \emph{Balanced Error Rate}, which we find to be a good proxy for the average precision:
\begin{equation}
 \Delta(Y, Y_c) = \frac{1}{2}\biggl[ \underbrace{\frac{|Y^{\text{pos}} \setminus Y_c^{\text{pos}}|}{|Y_c^{\text{pos}}|}}_{\text{false positive rate}} + \underbrace{\frac{|Y^{\text{neg}} \setminus Y_c^{\text{neg}}|}{|Y_c^{\text{neg}}|}}_{\text{false negative rate}} \biggr],
\label{eq:loss}
\end{equation}
where $Y^{\text{pos}}$ is shorthand for the set of images with positive labels ($Y^\text{neg}$ for negatively labeled images, similarly for $Y_c$). The Balanced Error Rate is designed to assign equal importance to false positives and false negatives, such that `trivial' predictions (all labels positive or all labels negative), or random predictions have loss $\Delta(Y, Y_c) = 0.5$ on average, while systematically incorrect predictions yield $\Delta(Y, Y_c) = 1$.

Other loss functions, such as the $0/1$ loss, could be optimized in our framework, though we find the loss of \eq{eq:loss} to be a better proxy for the average precision.

We optimize \eq{eq:objective} using the solver of \cite{bmrm}, which merely requires that we specify a loss function $\Delta(Y, Y_c)$, and procedures to solve \eq{eq:gmodel} and \eq{eq:colgen}. The solver must be modified to ensure that $\theta^\text{edge}_c$ remains positive. A similar modification was suggested in \cite{PetCae11}, where it was also used to ensure supermodularity of an optimization problem similar to that of \eq{eq:gmodel}.

\section{Experiments}
\label{sec:experiments}

We study the use of social metadata for three binary classification problems: predicting image labels, tags, and groups. Note some differences between these three types of data: labels are provided by human annotators outside of Flickr, who provide annotations based purely on image content. Tags are less structured, can be provided by any number of annotators, and can include information that is difficult to detect from content alone, such as the camera brand and the photo's location. Groups are similar to tags, with the difference that the groups to which a photo is submitted are chosen entirely by the image's author.

\xhdr{Data setup}
As described in Section \ref{sec:model}, for our first-order/node features $\phi_c(x_i)$ we construct indicator vectors encoding those words, groups, and tags that appear in the image $x_i$. We consider the 1000 most popular words, groups, and tags across the entire dataset, as well as any words, groups, and tags that occur at least twice as frequently in positively labeled images compared to the overall rate (we make this determination using only \emph{training} images). As word features we use text from the image's title, description, and its comment thread, after eliminating stopwords.

For our relational/edge features $\phi_c(x_i, x_j)$ we consider seven properties:
\begin{itemize*}
 \item The number of common tags, groups, collections, and galleries
 \item An indicator for whether both photos were taken in the same location (GPS coordinates are organized into distinct `localities' by Flickr)
 \item An indicator for whether both photos were taken by the same user
 \item An indicator for whether both photos were taken by contacts/friends
\end{itemize*}

Where possible, we use the training/test splits from the original datasets, though in cases where test data is not available, we form new splits using subsets of the available data. Even when the original splits are available, around 10\% of the images are discarded due to their metadata no longer being available via the Flickr API. This should be noted when we report results from other's work.

\xhdr{Evaluation}
Where possible we report results directly from published materials on each benchmark, and from the associated competition webpages. We also report the performance obtained using image tags alone (the most common form of metadata used by multimodal approaches), and a `flat' model that uses an indicator vector to encode collections, galleries, locations, and users, and is trained using an SVM; the goal of the latter model is to assess the improvement that can be obtained by using metadata, but not explicitly modeling \emph{relationships} between images. To report the performance of `standard' low-level image models we computed 1024-dimensional features using the publicly-available code of \cite{koen}; although these features fall short of the best performance reported in competitions, they are to our knowledge state-of-the-art in terms of publicly available implementations.

We report the Mean Average Precision (MAP) for the sake of comparison with published materials and competition results. For this we adopt an approach commonly used for SVMs, whereby we rank positively labeled images followed by negatively labeled images according to their first-order score $\langle \phi_c(x_i), \theta^\text{node}_c \rangle$. We also report performance in terms of the Balanced Error Rate $\Delta$ (or rather, $1~-~\Delta$ so that higher scores correspond to better performance).

\subsection{Image Labeling}
\label{sec:labeling}

Figure \ref{fig:plot_average} (left) shows the average performance on the problem of predicting image labels on our four benchmarks. We plot the performance of the tag-only flat model, all-features flat model and our all-features graphical model.

For ImageCLEF, the graphical model gives an 11\% improvement in Mean Average Precision (MAP) over the tag-only flat model, and a 31\% improvement over the all-features flat model. Comparing our method to the best text-only method reported in the ImageCLEF 2011 competition \cite{nowak10}, we observe a 7\% improvement in MAP. Our method (which uses no image features) achieves similar performance to the best visual-only method. Even though the images were labeled by external evaluators solely based on their \emph{content}, it appears that the social-network data contains information comparable to that of the images themselves. We also note that our graphical model outperforms the best visual-only method for 33 out of 99 categories, and the flat model on all but 9 categories.

On the PASCAL dataset we find that the graphical model outperforms the tag-only flat model by 71\% and the all-features flat model by 19\%. The performance of our model on the PASCAL dataset falls short of the best visual-only methods from the PASCAL competition; this is not surprising, since photos in the dataset have by far the least metadata, as discussed in Section 2 (Table \ref{tab:stats}).

On the MIR dataset the graphical model outperforms the tag-only and all-features flat models by 38\% and 19\%, respectively. Our approach also compares favorably to the baselines reported in \cite{huiskes10}. We observe a 42\% improvement in MAP and achieve better performance on all 14 categories except `night'.

On the NUS dataset our approach gives an approximately threefold improvement over our baseline image features. While the graphical model only slightly outperforms the tag-only flat model (by 5\%), we attribute this to the fact that some edges in NUS were suppressed from the graph to ensure that the model could be contained in memory. We also trained SVM models for six baseline features included as part of the NUS dataset \cite{nus-wide}, though we report results using the features of \cite{koen}, which we found to give the best overall performance.

Overall, we note that in terms of the Balanced Error Rate $\Delta$ the all-features flat model reduces the error over the tag-only model by 18\% on average (the all-features flat model does not fit in memory for the NUS data), and the graphical model performs better still, yielding a 32\% average improvement over the tag-only model. In some cases the flat model exhibits relatively good performance, though upon inspection we discover that its high accuracy is primarily due to the use of words, groups, and tags, with the remaining features having little influence. Our graphical model is able to extract additional benefit for an overall reduction in loss of 17\% over the all-features flat model. Also note that our performance measure is a good proxy for the average precision, with decreases in loss corresponding to increases in average precision in all but a few cases.

Although we experimented with simple methods for combining visual features and metadata, in our experience this did not further improve the results of our best metadata-only approaches.

\subsection{Tag and Group Recommendation}
\label{sec:recommendation}

We can also adapt our model to the problem of suggesting tags and groups for an image, simply by treating them in the same way we treated labels in Section \ref{sec:labeling}. One difference is that for tags and groups we only have `positive' groundtruth, i.e., we only observe whether an image \emph{wasn't} assigned a particular tag or submitted to a certain group, not whether it \emph{couldn't} have been. Nevertheless, our goal is still to retrieve as many positive examples as possible, while minimizing the number of negative examples that are retrieved, as in \eq{eq:loss}. We use the same features as in the previous section, though naturally when predicting tags we eliminate tag information from the model (sim.~for groups).

\begin{figure}[t]
 \begin{center}
  \includegraphics[scale=0.4]{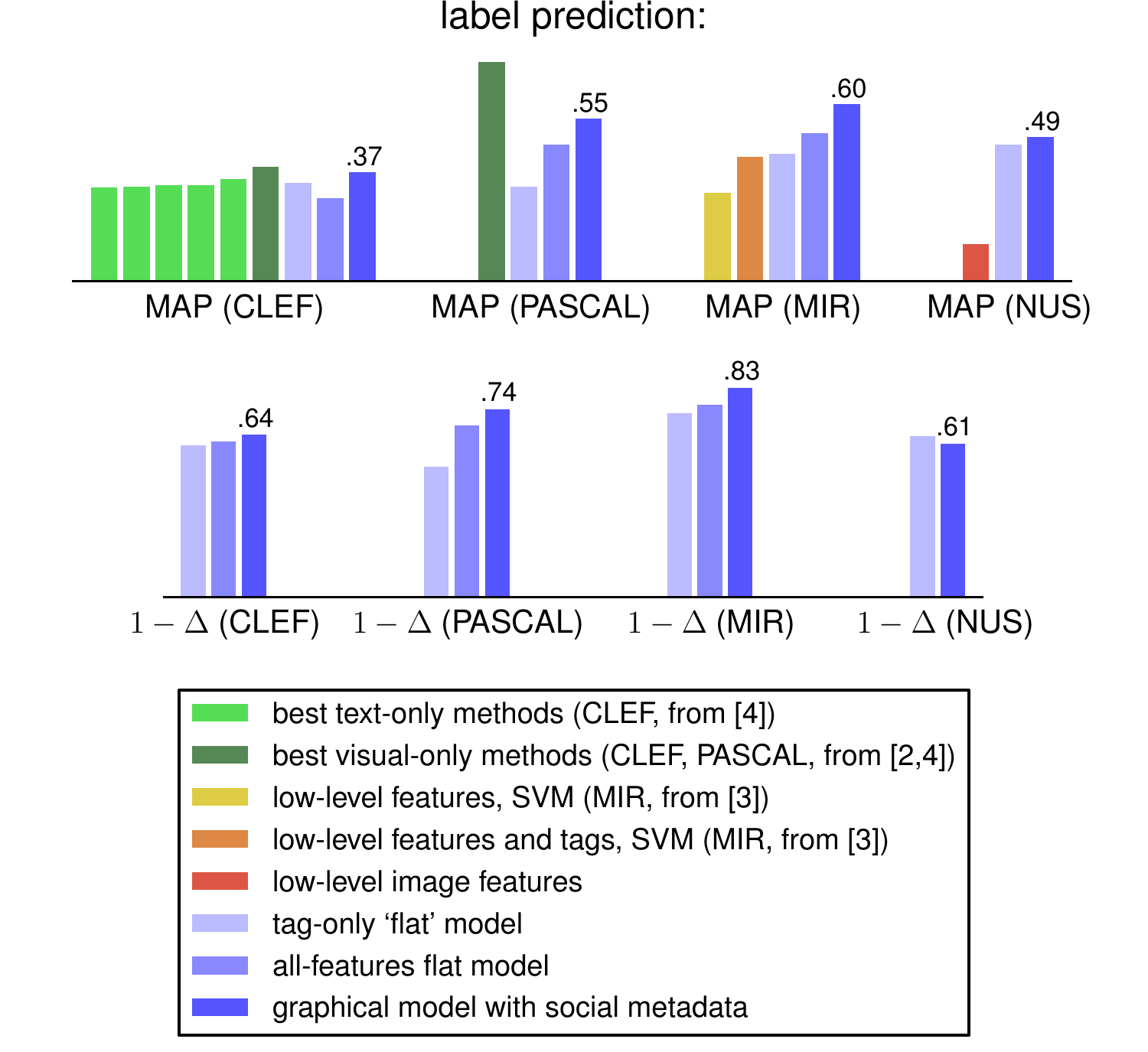}\hspace{-1mm}\includegraphics[scale=0.4]{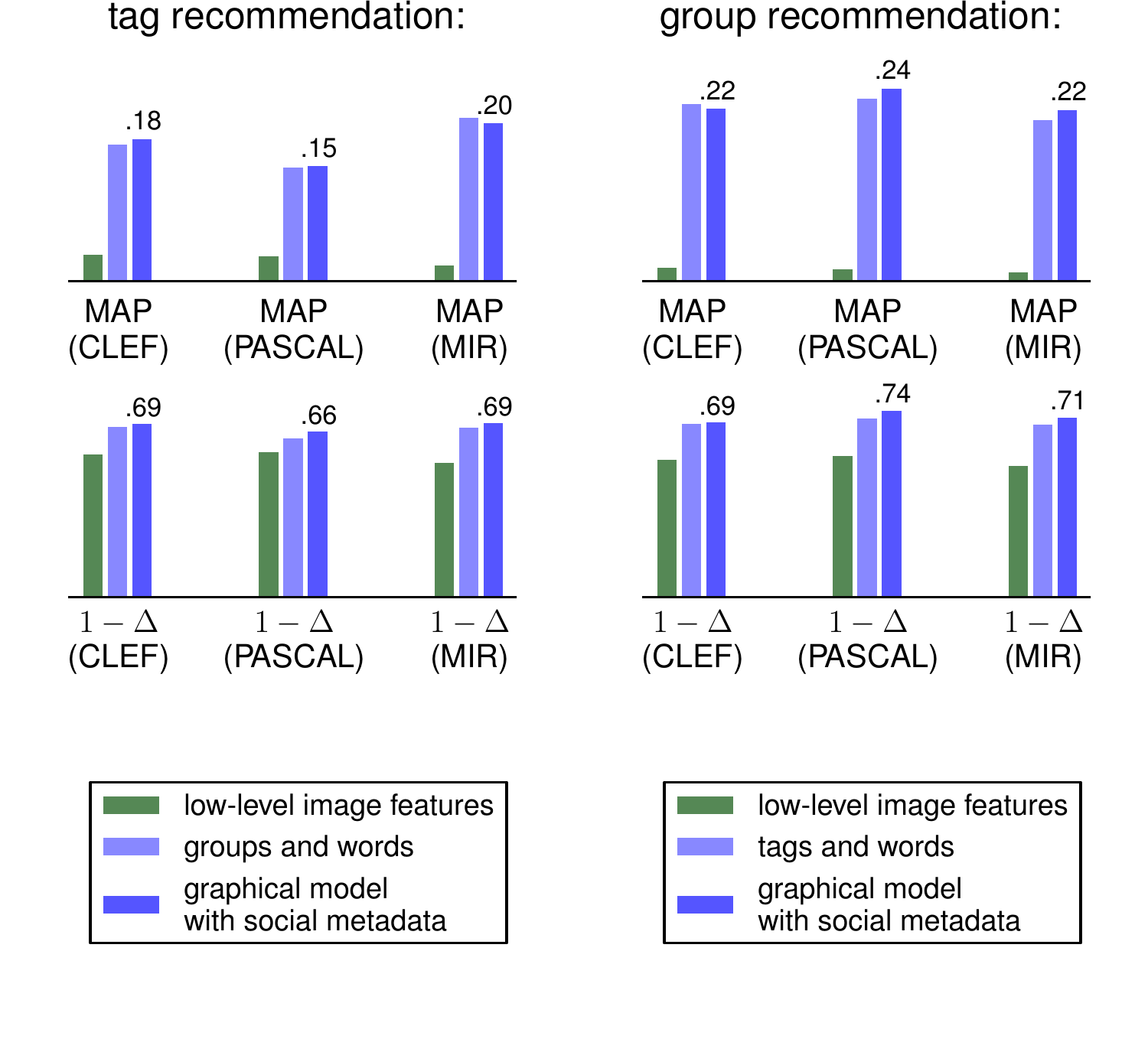}

\vspace{-3mm}
 \end{center}
\caption{Results in terms of the Mean Average Precision (top), and the Balanced Error Rate (bottom). `Flat' models use indicator vectors for all relational features and are trained using an SVM. Recall that using our performance measure, a score of $0.5$ is no better than random guessing. Comparisons for the ImageCLEF and PASCAL datasets are taken directly from their respective competition webpages; SVM comparisons for the MIR dataset are taken directly from \cite{huiskes10}.}
\label{fig:plot_average}
\end{figure}

Figure \ref{fig:plot_average} (center and right) shows the average performance of our model on the 100 most popular tags and groups that appear in the ImageCLEF, PASCAL, and MIR datasets. Using tags, groups, and words in a flat model already significantly outperforms models that use only image features; in terms of the Balanced Error Rate $\Delta$, a small additional benefit is obtained by using relational features.

\begin{figure*}[t]
 \begin{center}
  \includegraphics[scale=0.4]{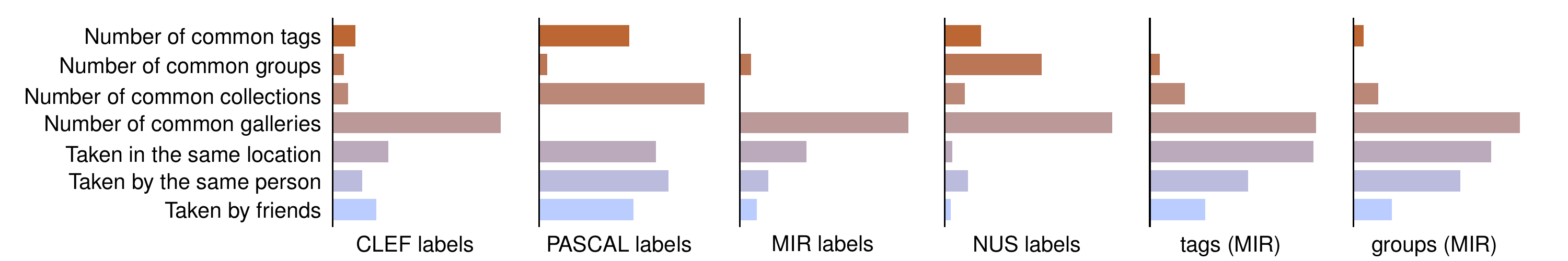}

\vspace{-3mm}
 \end{center}
\caption{Relative importance of social features when predicting labels for all four datasets, and groups, and tags on the MIR dataset (weight vectors for tags and groups on the remaining datasets are similar). Vectors were first normalized to have unit sum before averaging, as the models are scale-invariant.}
\label{fig:plot_clef_weights}
\end{figure*}

While image labels are biased towards categories that can be predicted from image contents (due to the process via which groundtruth is obtained), a variety of popular groups and tags can be predicted much more accurately by using various types of metadata. For example, it is unlikely that one could determine whether an image is a picture of the uploader based purely on image contents, as evidenced by the poor performance of image features the `selfportrait' tag; using metadata we are able to make this determination with high accuracy. Many of the poorly predicted tags and groups correspond to properties of the camera used (`50mm', `canon', `nikon', etc.). Such labels could presumably be predicted from exif data, which while available from Flickr is not included in our model.

\subsection{Social-Network Feature Importance}

Finally we examine which types of metadata are important for the classification tasks we considered. Average weight vectors for the relational features are shown in Figure \ref{fig:plot_clef_weights}. Note that different types of relational features are important for different datasets, due to the varied nature of the groundtruth labels across datasets. We find that shared membership to a gallery is one of the strongest predictors for shared labels/tags/groups, except on the PASCAL dataset, which as we noted in Section \ref{sec:datasets} was mostly collected before galleries were introduced in Flickr. For tag and group prediction, relational features based on location and user information are also important. Location is important as many tags and groups are organized around geographic locations. For users, this phenomenon can be explained by the fact that unlike labels, tags and groups are \emph{subjective}, in the sense that individual users may tag images in different ways, and choose to submit their images to different groups.

\small

\xhdr{Acknowledgements}
We thank the creators of each of the datasets used in our study for providing Flickr image sources. We also thank Jaewon Yang and Thomas Mensink for proofreading and discussions.
This research has been supported in part by NSF CNS-1010921, IIS-1016909, IIS-1159679, CAREER IIS-1149837, AFRL FA8650-10-C-7058, Albert Yu \& Mary Bechmann Foundation, Boeing, Allyes, Samsung, Yahoo, Alfred P. Sloan Fellowship and the Microsoft Faculty Fellowship.

\end{document}